%% file: sample-sigconf.tex
\documentclass[sigconf,natbib=false,nonacm,9pt]{acmart}
\AtBeginDocument{%
  }

\setcopyright{acmlicensed}
\copyrightyear{2025}
\acmYear{2025}
\acmDOI{XXXXXXX.XXXXXXX}
\usepackage{cite}
\usepackage{amsmath,amsfonts}

\usepackage[noend]{algpseudocode}
\usepackage{algorithmicx}
\usepackage{algorithm}
\usepackage{graphicx}
\usepackage{textcomp}
\usepackage{stfloats}

\usepackage{xcolor}
\usepackage[table]{xcolor}

\usepackage{booktabs}
\usepackage{enumitem}
\usepackage{listings}
\usepackage{hyperref} 
\usepackage{mathtools}
\usepackage{wrapfig}

\usepackage{tikz}
\usetikzlibrary{arrows.meta,positioning,shapes.geometric,calc}

\usepackage{pifont}
\newcommand{\cmark}{\textcolor{green!60!black}{\ding{51}}}
\newcommand{\xmark}{\textcolor{red}{\ding{55}}} 
\usepackage{setspace}
\usepackage[font=footnotesize,skip=0pt]{caption}
\usepackage{threeparttable}
\setlist[enumerate]{%
    leftmargin=0pt,
}
\begin{document}

%%
%% The "title" command has an optional parameter,
%% allowing the author to define a "short title" to be used in page headers.
\title{Scale-Gest: Scalable Model-Space Synthesis and Runtime Selection for On-Device Gesture Detection}

%%
%% The "author" command and its associated commands are used to define
%% the authors and their affiliations.
%% Of note is the shared affiliation of the first two authors, and the
%% "authornote" and "authornotemark" commands
%% used to denote shared contribution to the research.
% \author{Ben Trovato}
% \authornote{Both authors contributed equally to this research.}
% \email{trovato@corporation.com}
% \orcid{1234-5678-9012}
% \author{G.K.M. Tobin}
% \authornotemark[1]
% \email{webmaster@marysville-ohio.com}
% \affiliation{%
%   \institution{Institute for Clarity in Documentation}
%   \city{Dublin}
%   \state{Ohio}
%   \country{USA}
% }

\author{Abdul Basit, Saim Rehman, Muhammad Shafique}
\affiliation{%
  \institution{New York University (NYU) Abu Dhabi, Abu Dhabi, UAE}
  \city{}
  \country{}}
\email{abdul.basit@nyu.edu, sr7849@nyu.edu, muhammad.shafique@nyu.edu}
% \author{Anonymous Authors}

%%
%% By default, the full list of authors will be used in the page
%% headers. Often, this list is too long, and will overlap
%% other information printed in the page headers. This command allows
%% the author to define a more concise list
%% of authors' names for this purpose.
% \renewcommand{\shortauthors}{Trovato et al.}

%% The abstract is a short summary of the work to be presented in the
%% article.
\input{Sections/1_abstract}

% \keywords{
% on-device gesture detection, energy-aware vision, scalable model families, runtime adaptation, ROI tracking, YOLO}
%% This command processes the author and affiliation and title
%% information and builds the first part of the formatted document.
\maketitle

\input{Sections/2_introduction}

\input{Sections/3_background}
\input{Sections/4_methodology}
\input{Sections/5_experimental_setup}
\input{Sections/6_results}

\input{Sections/7_conclusion}

\bibliographystyle{acm}
\bibliography{cite}

\end{document}

%% file: Sections/1_abstract.tex
\begin{abstract}
Realizing on-device ML-based gesture detection under tight real-time performance, energy and memory constraints is challenging, especially when considering mobile devices with varying battery-power levels. Existing EdgeAI deployments typically rely on a single fixed detector, limiting optimization opportunities. We present \textit{Scale-Gest}, a novel run-time adaptive gesture detection framework that expands the detector space into a dense family of tiny-YOLO architectures. \textit{We introduce multiple novel device-calibrated ACE (Accuracy-Complexity-Energy) profiles} by analyzing different model-resolution-stride operating points. A \textit{lightweight run-time controller} selects an appropriate ACE mode under user-defined and battery constraints, while a motion-aware hand-gesture-tracking ROI gate crops the input for reduced complexity detection. To evaluate performance of our system in real-world car driving scenarios, we introduce a temporally-annotated Driver Simulated Gesture (DSG-18) dataset. Scale-Gest maintains event-level F1 while significantly reducing energy and latency compared to single-detector approaches. On a battery-powered laptop running gesture streams, our ACE controller reduces per-frame energy by $\approx$4× (from $\approx$6.9 mJ to $\approx$1.6 mJ) while maintaining high gesture-detection performance (event-level F1 $\approx$ 0.8–0.9) and low mean latency ($\approx$6 ms).

\end{abstract}

% Across HaGRID and DSG-18, Scale-Gest maintains event-level F1 while reducing energy by up to X× and latency by Y% compared to single-detector baselines.

% \begin{IEEEkeywords}
% on-device gesture detection, energy-aware vision, scalable model families, run-time adaptation, ROI tracking, YOLO
% \end{IEEEkeywords}

%% file: Sections/2_introduction.tex
\vspace{-10pt}
\section{Introduction and Related Works}

Gesture-based interaction is increasingly used in vehicle infotainment controls, smart displays, IoT-based home appliances, and service robots, where on-device processing is preferred for cost, privacy, and responsiveness. These platforms operate under tight power and memory constraints, and often share compute resources with other on-board tasks. At the same time, human interaction is temporally sparse; naturalistic driving studies report only a handful of infotainment interactions per hour, typically occupying less than \textit{3\% of driving time}~\cite{Angell2019,Neurauter_Hankey_Young_2007}. This necessitates a fundamental trade-off  between efficiency and accuracy under varying scenarios. For instance, high-fidelity detectors are needed to reliably capture short, small, or low-contrast gestures, yet running such models at constant resolution and frame rate wastes energy on most frames. Conversely, naively fixing the system to a tiny deep learning model reduces computations, but it risks missing critical gestures, motivating a run-time adaptive pipeline that can switch between high- and low-fidelity configurations based on the run-time context.

% However, state-of-the-art edge vision frameworks still leave three gaps
\textbf{State-of-the-Art and their Limitations:}
Modern single-shot detectors and mobile backbones, from SSD and EfficientDet to the YOLO families~\cite{ssd,Redmon2018YOLOv3,yolov4,Ge2021YOLOX,Jocher2023Ultralytics,yolov12}, offer multiple model scales and feature heads. Efficient inference techniques such as quantization and pruning~\cite{Han2015DeepCompression,Jacob2018Quantization,Wang_2021_CVPR,ahn2023safpyolo,hu2021microyolo} further reduce compute and memory requirements. Edge-AI schedulers and cross-layer frameworks~\cite{NestDNN,FlexDNN,SpecialSession,fi12070113,ROMANet,PENDRAM,Younesi2024ACS,s25061687,thermalmgmt} reason about resource budgets, performance isolation, and reliability across applications. These techniques, however, primarily act as \emph{offline} design-time optimizations, i.e, once a compressed model is selected, its operation and cost are largely fixed at run time, it does not adapt to varying battery states, thermal headrooms, or workload sparsity.

Early-exit architectures such as BranchyNet~\cite{BranchyNet}, SkipNet~\cite{Wu2018SkipNet} and FlexDNN~\cite{FlexDNN} adapt computation to input difficulty by adding auxiliary classifiers or learned gates that skip layers or exit early when confidence is high. However, the additional exit branches introduce overhead and can cumulate latency on hard inputs that must traverse multiple exits. Moreover, the scenario-dependent control to determine the early exiting is also not explored, as we do in this paper. NestDNN~\cite{NestDNN} instead builds a single multi-capacity network via nested descendant models and uses a run-time scheduler to pick a variant based on monitored system resources, but it remains resource-agnostic at the immediate input level and does not react to event-level temporal sparsity or short-lived gesture bursts. None of these state-of-the-art approaches react to dynamically varying battery-levels, video content, and user-defined constraints.

Video-specific methods like Deep Feature Flow~\cite{Li2016DeepFeatureFlow} reuse features across frames, while AdaScale~\cite{AdaScale} predicts an image-specific resize factor using the current frame to set the scale for the next frame, trading spatial resolution for speed and accuracy. DyRA~\cite{dyra} instead learns an image-specific scale factor applied directly to the current input to enhance scale robustness and detection accuracy. These techniques typically control a single knob (depth, resolution, or feature reuse) and often require learned controllers or policy training. Table~\ref{tab:related-compare} provides a high-level comparison of existing techniques with our Scale-Gest framework.

% Early-exit architectures such as BranchyNet~\cite{BranchyNet}, SkipNet~\cite{Wu2018SkipNet} and FlexDNN~\cite{FlexDNN} adapt computation to input difficulty by adding auxiliary classifiers or learned gates to skip layers or exit early when confidence is high, but the added exit branches introduce overhead and can cumulate latency on hard inputs that must traverse multiple exits. NestDNN~\cite{NestDNN} instead builds a single multi-capacity network via nested descendant models and uses a run-time scheduler to pick a variant based on monitored system resources, but it remains resource-agnostic at the immediate input level and does not react to event-level temporal sparsity or short-lived gesture bursts. 

% Video-specific methods like Deep Feature Flow~\cite{Li2016DeepFeatureFlow} reuse features across frames, while AdaScale~\cite{AdaScale} predicts an image-specific resize factor using the current frame to optimize the scale for the next frame in the sequence, trading spatial resolution for speed and accuracy. In contrast, DyRA~\cite{dyra} learns an image-specific scale factor applied directly to the current input image to enhance scale robustness and detection accuracy. These techniques typically operate a single knob (depth, resolution, or feature reuse) and often require learned controllers or policy training. 

\vspace{-10pt}

\begin{table}[ht]
  \centering
  \caption{Comparison of adaptive frameworks with Scale-Gest.}
  % \vspace{-5pt}
  \label{tab:related-compare}
  \small
  \renewcommand{\arraystretch}{1.1}
  % \begin{threeparttable}
  \resizebox{\columnwidth}{!}{%
  \begin{tabular}{lcccccc}
    \toprule
    \textbf{Framework} & \textbf{Dyn.\ Res.} & \textbf{Dyn.\ Stride} & \textbf{Model} & \textbf{Energy-} & \textbf{Event-} & \textbf{Gesture} \\
                       &                     & \textbf{(FPS)}        & \textbf{complex.} & \textbf{aware} & \textbf{temp.} & \textbf{task} \\
    \midrule
    BranchyNet~\cite{BranchyNet}                   & \xmark & \xmark & \xmark & \xmark & \xmark & \xmark \\
    SkipNet~\cite{Wu2018SkipNet}                   & \xmark & \xmark & \xmark & \xmark & \xmark & \xmark \\
    Deep Feature Flow~\cite{Li2016DeepFeatureFlow} & \xmark & \cmark & \xmark & \xmark & \xmark & \xmark \\
    AdaScale~\cite{AdaScale}                       & \cmark & \xmark & \xmark & \xmark & \xmark & \xmark \\
    DyRA~\cite{dyra}                               & \cmark & \xmark & \xmark & \xmark & \xmark & \xmark \\
    NestDNN~\cite{NestDNN}                         & \xmark & \xmark & \cmark & \cmark & \xmark & \xmark \\
    FlexDNN~\cite{FlexDNN}                         & \xmark & \xmark & \cmark & \cmark & \cmark & \xmark \\
    Cross-layer works~\cite{SpecialSession,fi12070113} & \textit{var.} & \textit{var.} & \textit{var.} & \cmark & \xmark & \xmark \\
    \midrule
    Our work (Scale-Gest)                         & \cmark & \cmark & \cmark & \cmark & \cmark & \cmark \\
    \bottomrule
  \end{tabular}%
  }
  \begin{tablenotes}
    \footnotesize
    % \centering
    \item Dyn.\ Res.: dynamic spatial scaling. Dyn.\ Stride (FPS): dynamic temporal sampling.
    Model complex.: model complexity scaling. Energy-aware: uses energy-based metrics.
    Event-temp.: event-level temporal metrics. Gesture task: gesture detection focus.
    \cmark: feature supported; \xmark: not addressed. var.: capability varies across works.
  \end{tablenotes}
  % \end{threeparttable}
\end{table}

\vspace{-10pt}
\textbf{Problem Statement:}
% We address the challenge of meeting stringent latency and energy budgets across diverse devices and usage scenarios by systematically constructing and exploiting dense, device-aware Accuracy--Complexity--Energy (ACE) profiles for gesture detection, and combining them with ROI-based inference under a run-time controller.
To achieve robust, real-time gesture detection under dynamically varying compute budgets, battery levels, and video scene statistics, \textit{there is a need for a Run-Time Adaptive ML-based Gesture Detection Framework} that can leverage a wide range of energy, memory, performance, and accuracy trade-offs. However, enabling this requires the generation of multiple variants of a given ML-model with diverse accuracy vs. complexity tradeoffs, a  strategy to select an appropriate mode depending upon the run-time conditions, and a temporal evaluation benchmark. Towards this, we propose \textit{Scale-Gest}, an adaptive framework that integrates the novel concept of \textit{dense device-calibrated ACE (Accuracy-Complexity-Energy) profiles}, \textit{Kalman-gated ROI inference}, and \textit{an adaptive ACE profile selector} that continuously balances accuracy, throughput, and energy on the target device during the execution time.

% The central research challenge addressed in this work: systematically constructing and exploiting dense, device-aware \textbf{Accuracy--Complexity--Energy (ACE) modes} for gesture detection, and combining it with ROI-based inference so that a run-time controller can meet constrained latency and energy budgets across devices and usage scenarios.

\begin{figure*}[t]
  \centering
  \includegraphics[width=1\linewidth]{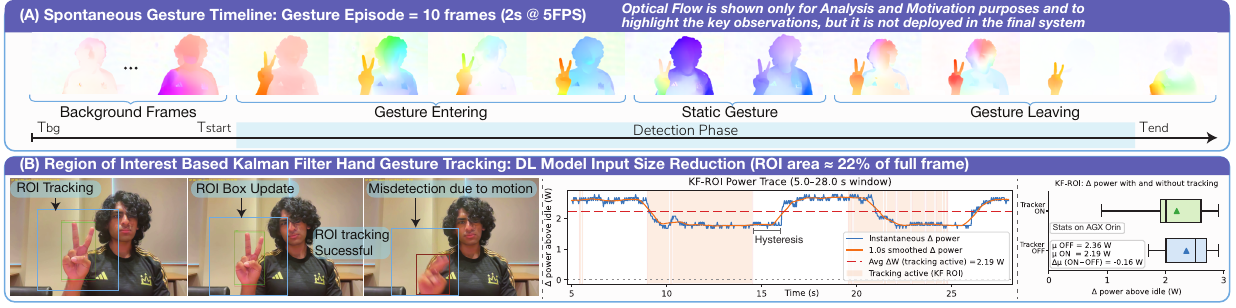}
    \caption{Temporal and spatial sparsity of driver gestures. (A) A typical DSG-18 gesture episode lasting only 10 frames, surrounded by long background periods, continuous high-fidelity inference is therefore wasteful. (B) A Kalman-based hand tracker crops the detector input to a small ROI, reducing effective input size. }
    \vspace{-10pt}
  \label{fig:gest_timeline_roi}
\end{figure*}

% while keeping the gesture clearly visible.
% \textit{Note: Optical Flow is only shown for analysis purpose, but not deployed in the Final System.}
% \textbf{Problem Statement:}
% \textit{This work addresses the challenge of meeting stringent latency and energy budgets across diverse devices and usage scenarios by systematically constructing and exploiting dense, device-aware \textbf{Accuracy--Complexity--Energy (ACE) modes} and combining them with ROI-based inference for robust gesture detection.
% }
% The central research challenge addressed in this work: systematically constructing and exploiting dense, device-aware \textbf{Accuracy--Complexity--Energy (ACE) modes} for gesture detection, and combining it with ROI-based inference so that a run-time controller can meet constrained latency and energy budgets across devices and usage scenarios.

\textbf{Our Motivational Analysis and Case Studies:}
A snapshot from \textit{our in-house generated \textbf{DSG-18} dataset} in Fig.~\ref{fig:gest_timeline_roi}A shows that naturalistic hand gestures occur as short, spontaneous episodes, while the surrounding stream is dominated by background frames. At the same time, gestures exhibit strong spatial sparsity, e.g., a driver’s hand usually occupies a small, trackable region in the video frame from the driver-facing view. Instead of compute-intensive motion tracking, we develop a\textit{ lightweight Kalman-based ROI tracker} that can keep the detector focused on this region instead of the full image (Fig.~\ref{fig:gest_timeline_roi}B), reducing the effective size of input feed to the ML-model without sacrificing gesture visibility. \textit{On an NVIDIA Jetson AGX Orin platform, this simple hand tracking reduces the instantaneous power above the idle baseline by about 0.16\,W on average for YOLOv12m, illustrating the potential energy savings of spatial gating}. Simple motion-based Region of Interest (ROI) gates~\cite{enBudget,7927209} can reduce spatial compute, but motion estimation at full resolution remains computationally and power demanding~\cite{Garca2013MultiGPUBO,10.1145/2228360.2228516}, hence un-affordable in such situations. 
% Therefore, many schemes rely on the detector to produce regions before sparsity can be exploited.

% A snapshot from our proposed DSG-18 dataset in Fig.~\ref{fig:gest_timeline_roi}A shows that naturalistic hand gestures occur as short, spontaneous episodes, while the surrounding stream is dominated by background frames. At the same time, gestures exhibit strong spatial sparsity. The driver’s hand usually occupies a small, trackable region of the driver-facing view, and a lightweight ROI tracker can keep the detector focused on this region instead of the full image (Fig.~\ref{fig:gest_timeline_roi}B). This suggests a path to reducing the effective input size without sacrificing visibility of the gesture. Our experiments results with YOLOv12m show that with hand tracking on, the instantaneous power above the idle baseline $\Delta$W reduces by 0.16W on average, Simple motion-based Region of Interest (ROI) gates can reduce spatial compute, but motion estimation at full resolution~\cite{enBudget,7927209} remains computationally and power demanding. Such schemes often rely on detector to produce regions before sparsity can be exploited.

\begin{figure}[ht]
  \centering
  \includegraphics[width=1\linewidth]{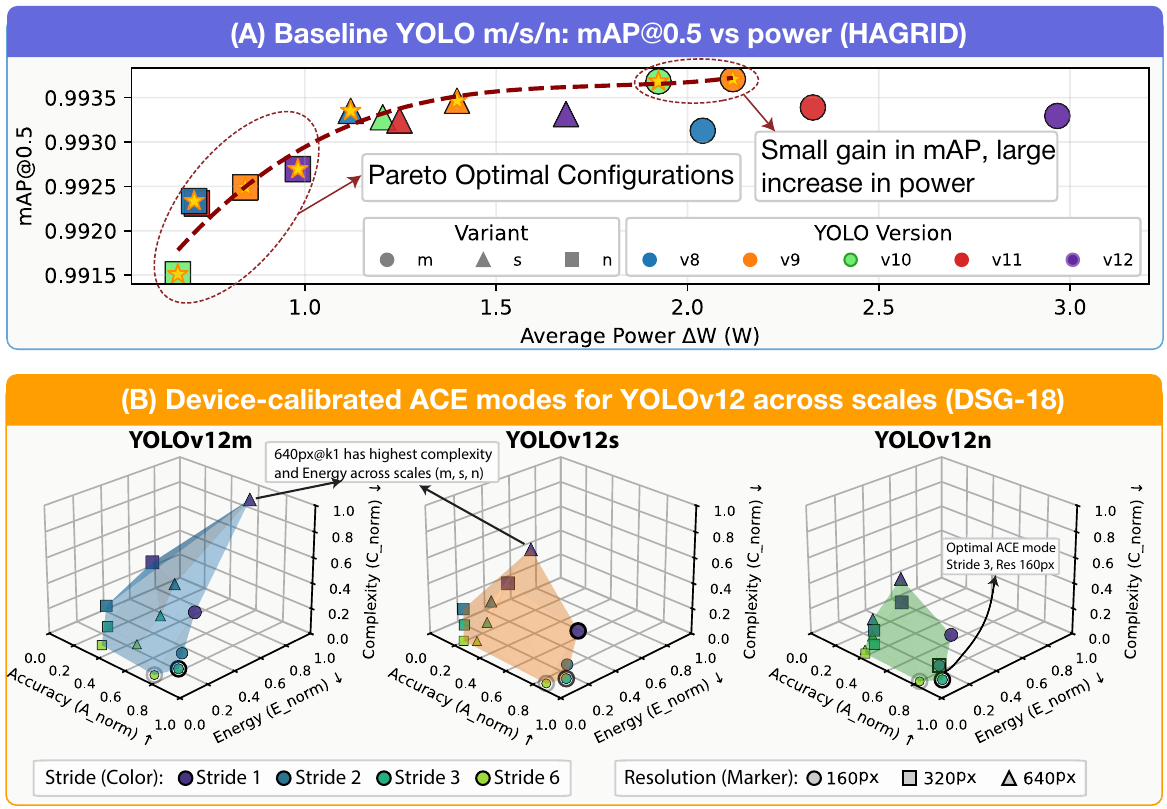}
   \caption{Device-calibrated Accuracy--Energy--Complexity (ACE) trade-offs. (A) Baseline YOLO m/s/n variants on a Jetson-class device show diminishing mAP@0.5 gains as average power increases on HaGRID. (B) Shows ACE profiles for YOLOv12 across scales on DSG-18 dataset by profiling different resolutions and strides, exposing several practical operating points.}    
  \label{fig:ace_modes}
\end{figure}

Moreover, device-level profiling reveals that an object detector's design space is much richer than a single configuration choice. On an NVIDIA Jetson AGX Orin platform, baseline detector variants show diminishing mAP@0.5 gains as average power increases (Fig.~\ref{fig:ace_modes}A). For a fixed backbone, profiling different resolutions and strides yields a dense set of device-calibrated ACE profiles, with multiple practical operating points beyond the standard model scales (Fig.~\ref{fig:ace_modes}B).

\vspace{-10pt}

% \vspace{-10pt}

% Moreover, device-level profiling reveals that the detector space is much richer than a single configuration choice. On a NVIDIA Jetson-class platform, baseline detector variants show diminishing mAP@0.5 gains as average power increases (Fig.~\ref{fig:ace_modes}(A)). Profiling different resolutions and strides yields a dense set of device-calibrated Accuracy--Complexity--Energy (ACE) modes, with multiple practical operating points beyond the standard model scales (Fig.~\ref{fig:ace_modes}B).

% . Running a high-fidelity detector at full frame rate on all inputs is therefore wasteful, even though this is the default choice in many deployed systems.
\vspace{-10pt}
\noindent\textit{\underline{Design Hint 1:}} If gesture episodes can be detected at event-level, high-cost ACE profiles only need to be activated around these short bursts, allowing the system to fall back to cheaper profiles during long idle periods and thus exploit temporal sparsity.

\noindent\textit{\underline{Design Hint 2:}} If a robust hand-tracking ROI gate keeps the detector input tightly cropped around the gesture region, the same backbone can operate at lower spatial complexity, reducing energy and latency while preserving detection quality.

\noindent\textit{\underline{Design Hint 3:}} If we treat each profiled configuration (backbone scale, resolution, and frame stride) as a point in a device-calibrated ACE profiles, a run-time controller can dynamically select operating modes according to accuracy, latency, and battery constraints, rather than committing to a single fixed detector configuration.

\noindent\textbf{Summarizing:}
These observations/hints point to three inter-related research challenges: (i) exploiting event-level temporal sparsity to time the activation of high-ACE profiles; (ii) leveraging spatial sparsity through ROI-based inference to reduce the effective input size; and (iii) systematically constructing and exploiting dense, device-aware ACE profiles so that a run-time controller can meet user-defined constraints and energy budgets across different devices and operational scenarios.

\textbf{Our Novel Contributions:}
Leveraging our above-discussed detailed analysis and design hints, we propose \emph{Scale-Gest (Section~\ref{sec:method}):}, an adaptive framework where the model space, device profiling, and spatial gating are treated as explicit optimization knobs under dynamic scenarios. Our key contributions are:

\begin{enumerate}
  \item \textbf{ACE Model-Space Synthesis (Section~\ref{subsec:model_space_Synthesis}):} We introduce a model-space synthesis method that turns standard detector configurations into a rich family of variants spanning depth/width multipliers, channel caps, and detection-head sets, while preserving graph correctness and deployment compatibility.
  \item \textbf{Device-Calibrated ACE Profiling and DSG-18 Dataset (Section~\ref{subsec:datasets_ace},~\ref{subsec:ace-profiling}):} We present a training and device-calibrated ACE profiling pipeline that measures frame- and event-level accuracy, stride-aware latency, and idle-normalized energy for each configuration on the target device. To investigate it in detail, we also introduce the \textit{temporally annotated Driver Simulated Gesture (DSG-18) dataset} with 18 classes for realistic benchmarking.
  \item \textbf{Constraint- and Context-aware run-time Selector (Section~\ref{subsec:adaptive_selector}):} We propose a run-time selector that operates on the ACE profiles to choose operating tiers under explicit accuracy, FPS, and system telemetry feedback, with stability ensured through windowed hysteresis and temporal smoothing.
  \item \textbf{ROI-based Spatial Gating (Section~\ref{subsec:kf-roi}):} We introduce a lightweight Kalman filter–based hand ROI tracker that propagates detector boxes over time to co-locate likely hand regions, enabling cropped inference that reduces compute and power while preserving detection quality.
\end{enumerate}

%% file: Sections/4_methodology.tex
\vspace{-10pt}
\section{Scale-Gest Framework}
\label{sec:method}

\begin{figure}[ht]
  \centering
  \includegraphics[width=\linewidth]{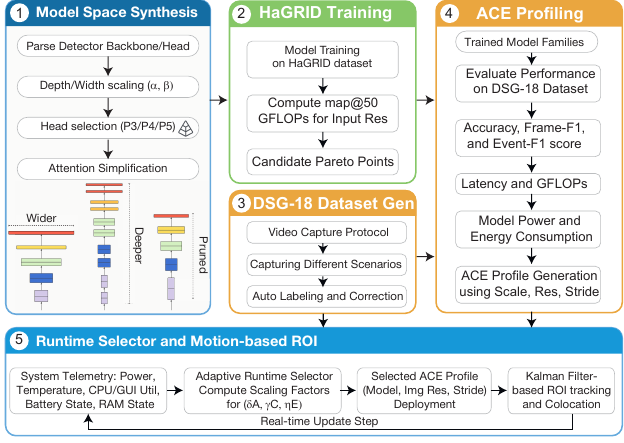}
  \caption{Scale-Gest Methodology: Outlines Components to realize the run-time adaptive framework. After Model Space Sampling, the training of each model family is performed on the HaGRID dataset, and similarly for run-time evaluation the DSG-18 dataset is curated and benchmarked to generate the model ACE profiles, which are used by the run-time selector for dynamically changing models during run time. }
  \label{fig:methodology}
\end{figure}

Scale-Gest combines an offline construction of device-calibrated ACE profiles with a run-time controller and ROI-based spatial gating. Our pipeline (1) expands the detector model space into multiple scales via graph-safe configuration synthesis, (2) trains these families on a large-scale hand-gesture corpus, (3) Curates DSG-18 Dataset for temporal profiling for gesture detection task, (4) profiles each model–resolution–stride operating point on the target SoC to build a device-calibrated ACE profile, and (5) selects operating profiles online under user and system constraints, aided by a motion-driven ROI spatial gate. Fig.~\ref{fig:methodology} summarizes the concept.
\vspace{-8pt}

% Scale-Gest is a run-time-adaptive, energy-constrained gesture detection framework that addresses the above gaps

% All accuracy, latency, and energy results are evaluated on DSG-18, a temporally annotated 18-class driver-gesture dataset that covers diverse lighting, hands, distances, and postures, providing a realistic target domain for event-level analysis.
% \label{subsec:overview}
% We develop a \textbf{run-time-adaptive, energy-constrained gesture detection} pipeline that (i) \emph{expands} the detector \emph{model-space} into \textbf{20 architecture families} across YOLOv8–v12 via graph-safe YAML synthesis; (ii) \emph{trains} every family \emph{from scratch on the full HaGRID} image dataset; (iii) \emph{profiles} each \emph{model$\times$resolution$\times$stride} operating point on the target SoC to build a device-calibrated \textbf{ACE table} (Accuracy–Complexity–Energy); and (iv) \emph{selects} the operating tier online under user/battery constraints, aided by a \textbf{motion-driven ROI} gate for sparse computation. 
% Evaluation is conducted on \textbf{DSG-18}, our temporally annotated 18-class suite with varied lighting, handedness, distance, and posture.
% The controller yields ranked tiers $\mathrm{Q}_1\!\ldots\!\mathrm{Q}_n$ and enforces hysteresis/dwell for stability.

% ========== SUBSECTION: MODEL SPACE ==========
% \subsection{ACE profiles construction}
\vspace{-20pt}
\subsection{Detector Model Space Synthesis}
\label{subsec:model_space_Synthesis}
% We start from standard detector backbones and generate a family of detector variants by scaling depth and width and by selecting different subsets of detection heads (P3–P5). Each family is parameterized by depth/width multipliers, a maximum channel cap, and a head set, spanning compact single-head designs up to three-head variants of moderate capacity. A simple, graph-safe transformation of the original configuration file adjusts layer repeats and channels and prunes unused heads while preserving operator compatibility, so that all generated models remain deployable on the target SoC. 

Scale-Gest builds a dense detector model space by \textit{automatically} generating multiple model variants from a single base configuration. Rather than hand-designing each backbone, we define a family of architectures by three knobs: (i) depth and width multipliers $(\alpha,\beta)$ that scale layer repeats and channels, (ii) a maximum channel cap $C^{\max}$ that limits peak width, and (iii) a subset of detection heads $\mathcal{H} \subseteq \{P3,P4,P5\}$ that controls which feature levels are used for prediction. By choosing these parameters to span compact single-head designs up to moderate-capacity three-head designs, we obtain a spectrum of candidate backbones.

\begin{wrapfigure}[16]{r}{0.11\textwidth}
\vspace{-15pt}
\hspace{-20pt}
\begin{tikzpicture}[
  node distance = 2mm,
  every node/.style = {font=\scriptsize},
  io/.style   = {trapezium, trapezium left angle=70, trapezium right angle=110,
                 draw, fill=gray!15, inner sep=1pt, text width=22mm, rounded corners,
                 align=center},
  proc/.style = {rectangle, draw, fill=blue!10, rounded corners,
                 inner sep=1pt, text width=24mm, align=center},
  ->, >=Latex
]

\node[io]                  (start) {Base Detector Config};
\node[proc,below=of start] (parse) {Parse layers \& links\\Locate \texttt{Detect}, P3/P4/P5};
\node[proc,below=of parse] (spec)  {Read family\\$(\alpha_f,\beta_f,C_f^{\max},\mathcal{H}_f)$};
\node[proc,below=of spec]  (closure){Dependency closure\\for \texttt{Detect} \& heads $\mathcal{H}_f$};
\node[proc,below=of closure] (scale){Scale repeats/channels\\by $(\alpha_f,\beta_f)$, cap $C_f^{\max}$};
\node[proc,below=of scale] (attn) {Simplify attention\\(optional)};
\node[proc,below=of attn] (reindex){Reindex links\\Split backbone/head};
\node[io,below=of reindex] (out) {Config for family $f$};

\draw (start)  edge (parse)
      (parse)  edge (spec)
      (spec)   edge (closure)
      (closure) edge (scale)
      (scale)  edge (attn)
      (attn)   edge (reindex)
      (reindex) edge (out);
\label{alg:modelspace}
\end{tikzpicture}

\caption{Model synthesis flowchart}
\label{flowchart}
\end{wrapfigure}
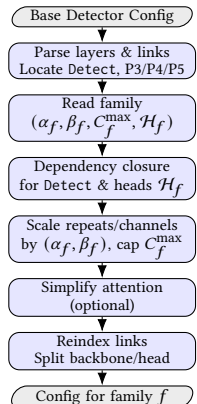

Given a standard detector configuration with \texttt{backbone} and \texttt{head} sections, we apply a graph-safe transformation to instantiate each family. For a chosen family, we compute the dependency closure of the \texttt{Detect} node and the heads in $\mathcal{H}$, retaining only the layers that are required to produce those features. This pruning preserves all necessary skip connections and intermediate feature maps while dropping unused branches.

On the retained subgraph, we scale the number of repeats and channels of each layer according to the family’s depth/width multipliers, and cap channels at $C^{\max}$ to bound model size. \textit{Channels are rounded to hardware-friendly granularity (e.g., multiples of 8) to maintain tensor-core efficiency.} For micro-scale families, heavy attention blocks in the backbone and head can optionally be replaced by lighter residual blocks to further reduce compute while preserving the overall graph topology. Finally, we re-index all \texttt{from} connections in the pruned graph and partition it back into \texttt{backbone} and \texttt{head} sections, yielding a deployable configuration for that family.
% Compact linear flowchart for model-space synthesis (single family f)
%
Flowchart in Fig.~\ref{flowchart} illustrate the process of detector model synthesis that forms the backbone dimension of our ACE profiles. Combining these with resolution and stride choices in later sections produces the full Accuracy--Complexity--Energy space explored by Scale-Gest.
% \begin{wrapfigure}[16]{r}{0.20\textwidth}
% \hspace{-20pt}

%   \centering
%   \includegraphics[width=1.10\linewidth]{Figures/Benchmark.pdf}
%   \caption{Accuracy--efficiency comparison of YOLOv8 and YOLOv10 synthesized families, showing how $C_{\max}$, depth ($\alpha$), width ($\beta$), and head count shape the accuracy-GFLOP trade-off.}
%   \label{fig:synth_models}
% \end{wrapfigure}

\begin{wrapfigure}[17]{l}{0.26\textwidth}
  \vspace{-13pt}                         
  
  \setlength{\intextsep}{2pt}       
  \setlength{\columnsep}{-30pt}      
  \centering
  \includegraphics[width=1.18\linewidth]{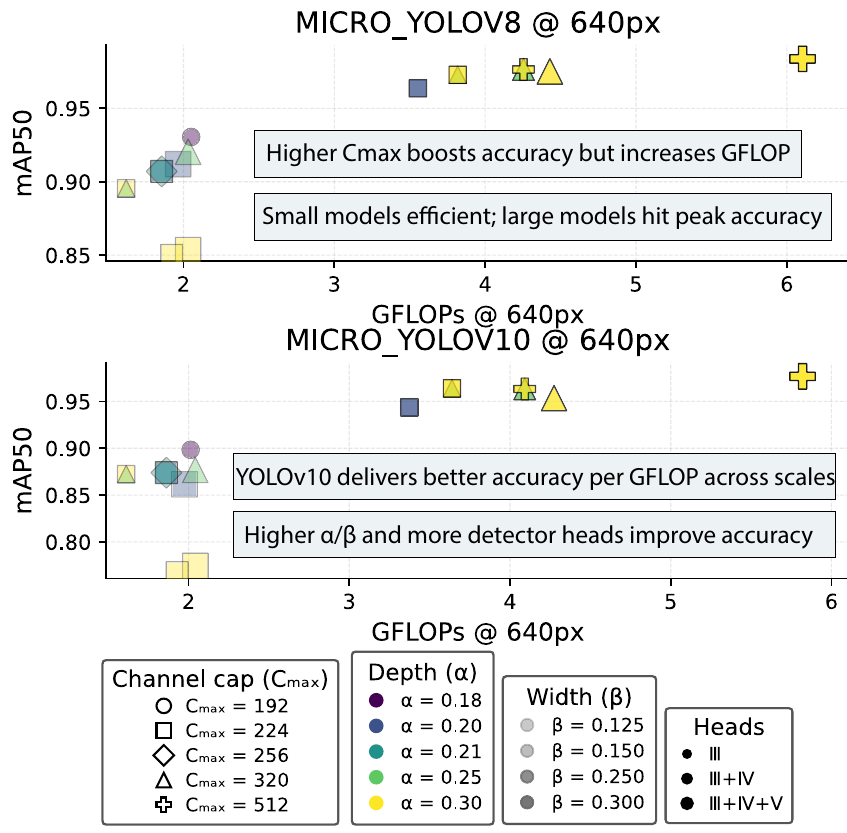}
  \caption{Accuracy--efficiency comparison of YOLOv8, v10 synthesized families.}
  \label{fig:synth_models}
\end{wrapfigure}

As illustrated in Fig.~\ref{fig:synth_models}, these trends help visualize how the synthesized models are generated how their structural choices translate into downstream performance on the HaGRID dataset. Increasing $C^{\max}$, depth ($\alpha$), width ($\beta$), or head count systematically boosts accuracy at the cost of higher GFLOPs, while lighter configurations achieve strong efficiency with moderate accuracy.

\subsection{Datasets and ACE Profiling Setup}
\label{subsec:datasets_ace}

\textbf{Design-Time Source (HaGRID):}
We use the HaGRID dataset~\cite{hagrid} as the \emph{source} domain for training and design-time model evaluation. All YOLO families are trained on the full 18-class taxonomy for 50 epochs, and we record classic image-level metrics (mAP@0.5, parameter count, FLOPs at \{160, 320, 640\} px). This profile provides stable priors on detector quality.
% and allows us to remove clearly dominated families before expensive video profiling.

% \vspace{-5pt}
\begin{wrapfigure}[14]{l}{0.225\textwidth}
\vspace{-15pt}
\hspace{-16pt}
  \setlength{\intextsep}{2pt}       
  \setlength{\columnsep}{-30pt}      
  \centering
  \includegraphics[width=1.18\linewidth]{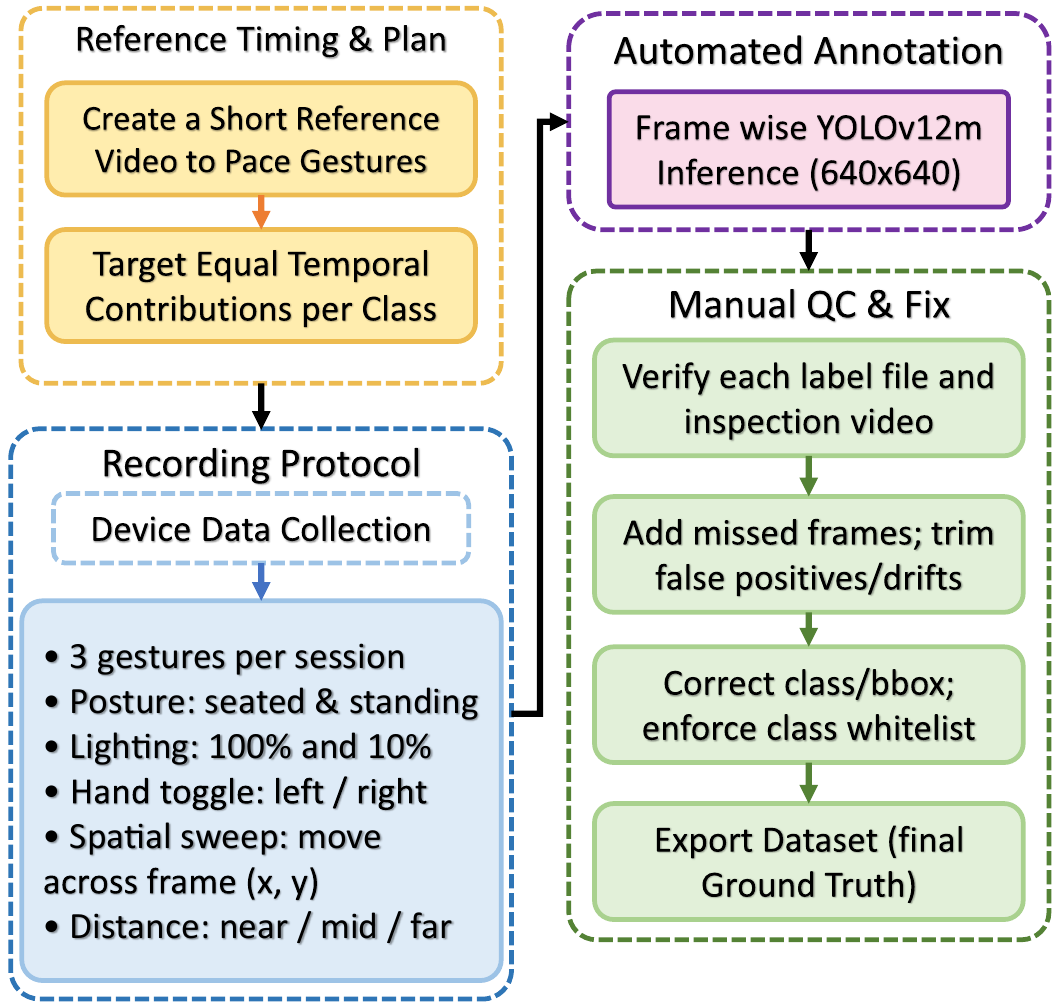}
  \caption{\textbf{DSG-18 dataset generation \& post-processing pipeline.}}
  \label{fig:DSG_collection}
\end{wrapfigure}
\vspace{-10pt}
\begin{figure}[ht]
  \centering
  \includegraphics[width=\linewidth]{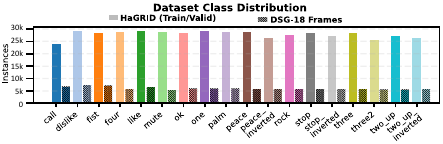}
  \caption{Dataset Class Distribution: Comparison of sample counts between HaGRID (Train/Valid) and total DSG-18 frames per gesture class.}
  \label{fig:acegest_dist}
\end{figure}

\begin{figure*}[b]
  \centering
  \includegraphics[width=\linewidth]{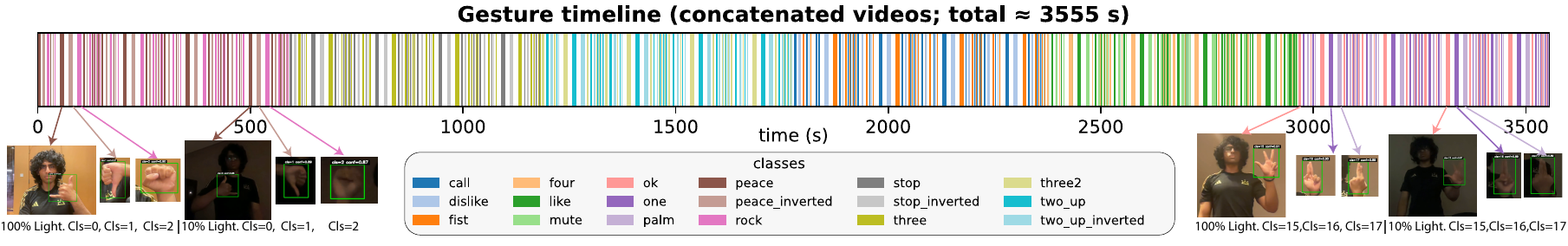}
  \caption{\textbf{Concatenated gesture timeline} for DSG-18 (\(\approx\)\,3555\,s).
  Color encodes gesture class. The sequence reveals temporally sparse and irregular gesture bursts.}
  \label{fig:acegest_timeline}
\end{figure*}

\vspace{-10pt}

\textbf{Run-Time Target (DSG-18):}
To capture temporal behavior under deployment-like conditions, we curate \emph{DSG-18}, a video suite aligned one-to-one with the HaGRID gesture classes. Each clip is recorded under a controlled but varied protocol detailed in Fig.~\ref{fig:DSG_collection}: triplets of gestures per session, seated and standing postures, high/low illumination, left/right hand toggling, spatial sweeps across the frame, and near/mid/far camera distances. Frames are first auto-annotated with a high-accuracy teacher detector trained on HaGRID dataset, and then manually validated to correct class IDs, bounding boxes, and occasional misses. Bechmarking on DSG-18 dataset enables evaluation of both frame- and event-level metrics. Class histograms and the concatenated gesture timeline confirm that (i) all 18 classes are represented and (ii) gesture episodes are short bursts embedded in background segments (Figs.~\ref{fig:acegest_dist}, \ref{fig:acegest_timeline}).

HaGRID and DSG-18 play distinct roles in our methodology. HaGRID is used only for training: it tells us which model families are competitive in terms of accuracy and nominal FLOPs, but it cannot characterize temporal knobs such as frame stride. ACE profiles for dynamic selection are therefore built exclusively on DSG-18. For each configuration (model family, input resolution, and frame stride) on the target SoC, we measure:
(i) frame-F1 and event-F1 on temporally annotated videos,
(ii) stride-aware latency statistics (mean and p90) as the complexity axis, and
(iii) idle-normalized power and energy per source frame as the energy axis.

\vspace{-8pt}

\subsection{Device–Calibrated ACE Profiling}
\label{subsec:ace-profiling}

% To expose realistic ACE trade–offs, we profile each trained detector configuration on the target hardware. A configuration is defined as \(x = (m, r, k)\), where $f$ is a detector family instance, $r\!\in\!\{160,320,640\}$ the input resolution, and $k\!\in\!\{1,2,3,6\}$ the frame stride.

\paragraph{Notation.}
A detector configuration is denoted by $x=(m,r,k)$, where 
$m\!\in\!\mathcal{M}$ is a trained model, 
$r\!\in\!\mathcal{R}$ the input resolution, and 
$k\!\in\!\mathcal{K}$ the frame stride. 
For each $(v,g)$ video/GT pair, the algorithm measures frame--F1 $A_{\text{fr}}$, event--F1 $A_{\text{ev}}$, effective latency $L_{\text{eff}}(x)$, optional FLOP cost $C_{\text{flop}}(x)$, instantaneous power $W(t)$, energy $E(x)$, and mean excess power $\overline{\Delta W}(x)$. 
% Aggregating over videos yields the accuracy score $A(x)$, compute/latency cost $C(x)$, and energy cost $E(x)$, which are normalized to $(\tilde{A}(x),\tilde{C}(x),\tilde{E}(x))$ via min--max scaling. 
The ACE score for configuration $x$ is
$
S_{\text{ACE}}(x)
= \delta_A\,\tilde{A}(x)
  - \gamma_C\,\tilde{C}(x)
  - \eta_E\,\tilde{E}(x),
$
where $(\delta_A,\gamma_C,\eta_E)$ are scaling factors controlling the accuracy, compute, and energy contributions. These scaling factors are set by the run-time selector (Sec.~\ref{subsec:adaptive_selector}). Algorithm~\ref{alg:ace-profile} details the process of ACE profling.

\begin{algorithm}[ht]
\caption{Device–calibrated ACE profiling}
\label{alg:ace-profile}
\footnotesize
\begin{algorithmic}[1]
\Require Trained models $\mathcal{M}$, resolutions $\mathcal{R}$, strides $\mathcal{K}$,
         DSG-18 videos with GT, device and power logger
\Ensure  ACE table with $(A,\tilde{A},\tilde{C},\tilde{E},S_{\text{ACE}})$ per configuration
\State $\mathcal{P} \gets \emptyset$ \Comment{list of raw profiles}
\For{each $m \in \mathcal{M}$}
  \For{each $r \in \mathcal{R}$}
    \For{each $k \in \mathcal{K}$}
      \State Reset accumulators for accuracy, latency, energy
      \For{each video / GT pair $(v,g)$}
        \State Run detector $m$ on $(v,g)$ at $(r,k)$ with hold–last imputation
        \State Measure frame F1 $A_{\text{fr}}$, event F1 $A_{\text{ev}}$
        \State Record per–call latency samples and power samples $W(t)$
        \State Accumulate per–video metrics
      \EndFor
      \State Aggregate over videos to get $A_{\text{fr}},A_{\text{ev}}$ and $A(x)$
      \State Compute $L_{\text{eff}}(x)$ and optional $C_{\text{flop}}(x)$
      \State Integrate idle–subtracted power to get $E(x)$ and $\overline{\Delta W}(x)$
      \State Append profile $p_x$ for configuration $x=(m,r,k)$ to $\mathcal{P}$
    \EndFor
  \EndFor
\EndFor
\State Min–max normalize $\{A(x)\}$, $\{L_{\text{eff}}(x),C_{\text{flop}}(x)\}$, and $\{E(x)\}$
       to obtain $(\tilde{A},\tilde{C},\tilde{E})$
\For{each profile $p_x \in \mathcal{P}$}
\State $S_{\text{ACE}}(x) \gets \delta_A\, \tilde{A}(x) - \gamma_C\, \tilde{C}(x) - \eta_E\, \tilde{E}(x)$
\EndFor
\State Save $\mathcal{P}$ (including raw metrics and $S_{\text{ACE}}$) as \texttt{ace\_profiles.json}
\end{algorithmic}
\end{algorithm}

\vspace{-10pt}
\paragraph{Accuracy:}
For each DSG-18 video, we run the detector on every $k$-th frame and apply \emph{hold–last} imputation on frames, which has greater than 1 stride, with an exponentially decaying  confidence. From frame-wise predictions and temporal annotations, we compute frame–level F1 ($A_{\text{fr}}$) and event–level F1 ($A_{\text{ev}}$), where a gesture event is counted as a true-positive if any predicted positive falls inside its time span.  We use a blended accuracy:
$
A(x) = \lambda A_{\text{ev}} + (1-\lambda) A_{\text{fr}},
$
with $\lambda\!\in\![0,1]$ fixed across experiments.

\paragraph{Complexity:}
We measure inference–only latency on the device after a short warm–up and record mean and percentiles. The stride –aware effective latency is
$
L_{\text{eff}}(x) = L_{\text{mean}}(x)/k
$
, which approximates per–source–frame cost at stride $k$. Optionally, we estimate effective GFLOPs at 640×640 and scale by $(r/640)^2/k$ to obtain $C_{\text{flop}}(x)$, used only for interpretability.
\vspace{-4pt}
\paragraph{Energy:}
We attach a lightweight power logger: \texttt{tegrastats} on NVIDIA Jetson and \texttt{nvidia-smi} on desktop GPUs. The logger samples instantaneous power $W(t)$ during execution. An idle baseline $W_{\text{idle}}$ is measured once per device, while system is running background tasks and averaged. For each configuration, we integrate the idle–subtracted trace,
\(
E(x) = \frac{1}{N_{\text{src}}}
       \int_{t_0}^{t_1} \max\!\big(W(t)-W_{\text{idle}}, 0\big)\,dt,
\)
yielding Joules-per-source frame and average excess power
$\overline{\Delta W}(x)$.
\vspace{-13pt}
\paragraph{Normalization and ACE Score:}
Across all configurations, we min –max normalize blended accuracy $A(x)$, effective latency and GFLOPs, and energy to obtain $(\tilde{A},\tilde{C},\tilde{E})\in[0,1]^3$.  Complexity $\tilde{C}$ is a weighted combination of normalized latency and FLOPs.

\vspace{-5pt}
\subsection{Adaptive Run-Time Selector}
\label{subsec:adaptive_selector}

Given the ACE profiles built on DSG-18 (Sec.~\ref{subsec:ace-profiling}), the \emph{Adaptive Run-Time Selector} dynamically chooses an appropriate detector configuration (i.e., model family, input resolution, stride, ROI policy) that best matches the current resource envelope of the device. The key idea is to keep the profiling stage fully offline, and to make the online selector lightweight. At run time, we only manipulate pre-computed ACE statistics and live telemetry, without re-running any retraining. 
\textbf{Inputs:}
The selector consumes: (i) a set of ACE profiles
\(\{p_i\}\), each containing blended accuracy \(A_i\) (event+frame F1), stride-normalized latency \(L_i\), effective FLOPs \(F_i\) and energy-per-source frame \(E_i\); (ii) application-level constraints, namely a minimum accuracy threshold \(A_{\min}\), a target frame rate \(\mathrm{FPS}_{\text{tgt}}\), and a battery model (battery capacity, state-of-charge, horizon, and background power); and (iii) live telemetry from the device, in form of CPU/GPU temperatures, GPU utilization, and battery percentage.

\textbf{Constraint Budgets and Slacks:}
From \(\mathrm{FPS}_{\text{tgt}}\), we derive a latency budget \(L_{\text{bud}} = 1/\mathrm{FPS}_{\text{tgt}}\) s per displayed frame. Given the battery and horizon parameters, we analytically convert the usable energy into a per-frame budget \(E_{\text{bud}}\) (or use a user-specified limit). We then mark each profile \(p_i\) as feasible if it satisfies \(A_i \ge A_{\min}\), \(L_i \le L_{\text{bud}}\) and \(E_i \le E_{\text{bud}}\). Over feasible profiles (or over all profiles if none are feasible) we compute \emph{slack} terms that summarize how comfortably the Pareto front sits inside each budget:
\(
s_{\text{lat}} = \frac{L_{\text{bud}} - \min_i L_i}{L_{\text{bud}}},\quad
s_{\text{energy}} = \frac{E_{\text{bud}} - \min_i E_i}{E_{\text{bud}}},\quad
s_{\text{acc}} = \frac{\max_i A_i - A_{\min}}{1 - A_{\min}},
\)
each clamped to \([0,1]\).
Intuitively, \(s_{\text{lat}}\approx 1\) means latency is easy to satisfy and can be deprioritized, while \(s_{\text{lat}}\approx 0\) means we are close to violating the frame-rate constraint.

\textbf{Telemetry Pressures:}
Live measurements are converted into dimensionless \emph{pressures}, i.e., thermal pressure (ratio of CPU/GPU temperature to a safe cap), utilization pressure (GPU utilization relative to a threshold), and battery pressure (residual fraction of capacity used). All are clamped in a small range (e.g., \([0,2]\) for thermal/utilization, \([0,1]\) for battery) and can be injected via the system log or explicit overrides. This lets the selector react sharply to overheating, GPU saturation or low battery, even when the average ACE profile suggests plenty of slack.

\textbf{Adaptive ACE Weights and Scoring:}
We combine slacks and pressures into raw weights for Accuracy, Complexity and
Energy:
% \[
% \begin{aligned}
% w_A^{\text{raw}} &\propto \exp(2\,s_{\text{acc}})\cdot \exp(1\!-\!\text{battery}),\\
% w_C^{\text{raw}} &\propto \exp(3(1\!-\!s_{\text{lat}}))\cdot \exp(2.5\,\text{thermal})\cdot \exp(1.5\,\text{util}),\\
% w_E^{\text{raw}} &\propto \exp(2.5(1\!-\!s_{\text{energy}}))\cdot \exp(3\,\text{battery}),
% \end{aligned}
% \]

\begingroup
    \vspace{-5pt}
    \setlength{\abovedisplayskip}{10pt} % spacing before
    \setlength{\belowdisplayskip}{10pt} % spacing after
    \small
    \[
    \begin{aligned}
        \delta_A^{\text{raw}} &\propto \exp(2\,s_{\text{acc}})\cdot \exp(1-\text{battery}),\\
        \gamma_C^{\text{raw}} &\propto \exp\bigl(3(1-s_{\text{lat}})\bigr)\cdot \exp(2.5\,\text{thermal})\cdot \exp(1.5\,\text{util}),\\
        \eta_E^{\text{raw}} &\propto \exp\bigl(2.5(1-s_{\text{energy}})\bigr)\cdot \exp(3\,\text{battery}),\\[4pt]
    \end{aligned}
    \]
\endgroup

then normalize them to $(\delta_A,\gamma_C,\eta_E)$ so that $\delta_A + \gamma_C + \eta_E = 1$.
In parallel, we normalize the ACE axes across all profiles:
\(\tilde{A}_i = \mathrm{minmax}(\{A_i\})\),
\(\tilde{E}_i = \mathrm{minmax}(\{E_i\})\), and
\(\tilde{C}_i\) as a mixture of latency and FLOPs:
\(
\tilde{C}_i = \alpha_{\text{lat}}\cdot \mathrm{minmax}(\{L_i\})_i + 
              \alpha_{\text{flop}}\cdot \mathrm{minmax}(\{F_i\})_i.
\)
To avoid double-counting, \(\alpha_{\text{flop}}\) is reduced when FLOPs and latency are strongly correlated. We estimate the Spearman correlation \(\rho(F,L)\) and set \(\alpha_{\text{flop}} = 0.3(1-|\rho|)\), \(\alpha_{\text{lat}} = 1-\alpha_{\text{flop}}\). Each profile then receives an adaptive ACE score:
$
S_i = \delta_A\,\tilde{A}_i \;-\; \gamma_C\,\tilde{C}_i \;-\; \eta_E\,\tilde{E}_i.
$
The selector ranks all profiles by \(S_i\); the top configuration is chosen as the active operating point, and the top-\(K\) are exposed for analysis.

\textbf{Real-Time Operation:}
In the experimental setup, we evaluate the selector in a closed loop on DSG-18. For each video, we read the latest telemetry sample, recompute $(\delta_A,\gamma_C,\eta_E)$ and \(\{S_i\}\), select the top-scoring configuration, and run the corresponding YOLO model with its profiled resolution and stride. The resulting latency, power and accuracy are logged alongside the chosen weights, demonstrating how the controller re-allocates priority between accuracy, throughput and energy as conditions change (e.g., battery drain or thermal throttling).
\vspace{-10pt}
\subsection{Kalman-Gated ROI Tracking}
\label{subsec:kf-roi}
Running the detector on the full input resolution stream is wasteful when the hand moves smoothly and occupies a small region. We therefore integrate a \emph{Kalman-gated region-of-interest (ROI) tracker} that steers the detector to a prediction-driven crop while preserving single-hand detection accuracy.

We maintain a single-object Kalman filter over the hand bounding box (center position and size). When no track is active, the detector $f_\theta$ is evaluated on the full frame and the highest-confidence hand box initializes the Kalman state. Once a track is active, each new frame is processed in three steps: (i) the Kalman filter predicts the next box $\hat{b}_t$, (ii) we construct a square ROI $R_t$ around $\hat{b}_t$ with side length $s\cdot\max(w(\hat{b}_t),h(\hat{b}_t))$, and (iii) we run $f_\theta$ \emph{only} on $I_t[R_t]$ and map any detection back to full-frame coordinates. The resulting box $b_t$ is accepted for tracking if its IoU with the prediction exceeds a gate $\tau$; otherwise the Kalman state is re-initialized from $b_t$.

% Track management is handled with a simple miss budget. If the ROI yields no valid detection for $T_{\text{miss}}$ consecutive frames, the tracker is dropped and the system falls back to full-frame detection until the hand is reacquired. This allows the pipeline to recover from occlusions or large motion jumps without accumulating error. In our implementation, each frame is annotated with a binary \texttt{track\_active} flag and instantaneous power $P_t$; post-processing the per-frame log produces $\Delta W(t)=P_t-P_{\text{idle}}$ traces and average $\Delta W$ over tracker-active intervals (cf. Fig.~\ref{fig:gest_timeline_roi}). Overall, the Kalman-gated ROI preserves frame-/event-level accuracy relative to full-frame evaluation while significantly reducing the average number of processed pixels and the corresponding energy per frame, making it a practical building block for on-device gesture detection under tight power budgets.

Track management uses a miss budget. If the ROI yields no valid detection for $T_{\text{miss}}$ consecutive frames, the tracker is dropped and the system falls back to full-frame detection until the hand is reacquired. This allows the pipeline to recover from occlusions or large motion jumps without accumulating error. In our implementation, each frame is annotated with a \texttt{track\_active} flag and power $P_t$; post-processing the per-frame log produces $\Delta W(t)=P_t-P_{\text{idle}}$ traces and average $\Delta W$ over tracker-active intervals (cf. Fig.~\ref{fig:gest_timeline_roi}). Overall, the Kalman-gated ROI preserves frame-/event-level accuracy relative to full-frame evaluation while reducing the average number of processed pixels and the corresponding energy per frame, making it a practical building block for on-device gesture detection under tight power budgets.

\vspace{-10pt}

\begin{algorithm}[ht]
\footnotesize
\caption{Kalman-gated ROI tracking for single-hand detection}
\label{alg:kalman-roi}
\begin{algorithmic}[1]
\Require Detector $f_\theta$, frames $\{I_t\}$, ROI scale $s$, IoU gate $\tau$, miss budget $T_{\text{miss}}$
\State $\mathcal{T}\gets\emptyset$ \Comment{no active track}, \ $m\gets0$
\While{streaming frames $I_t$}
  \If{$\mathcal{T}=\emptyset$} \Comment{(re)acquire on full frame}
    \State $b_t\gets f_\theta(I_t)$
    \If{$b_t\neq\varnothing$}
      \State initialize Kalman state from $b_t$; $\mathcal{T}\gets\text{active}$; $m\gets0$
    \EndIf
  \Else \Comment{track with ROI}
    \State $\hat{b}_t\gets\textsc{KalmanPredict}(\mathcal{T})$
    \State $R_t\gets$ square ROI around $\hat{b}_t$ with side $s\cdot\max(w(\hat{b}_t),h(\hat{b}_t))$
    \State $b_t^{\text{roi}}\gets f_\theta(I_t[R_t])$; map to full-frame box $b_t$ (if any)
    \If{$b_t=\varnothing$}
      \State $m\gets m+1$
      \If{$m\ge T_{\text{miss}}$} $\mathcal{T}\gets\emptyset$ \EndIf
    \Else
      \State $m\gets0$; $\text{IoU}\gets \mathrm{IoU}(b_t,\hat{b}_t)$
      \If{$\text{IoU}<\tau$}
        \State re-initialize Kalman state from $b_t$
      \Else
        \State \textsc{KalmanUpdate}$(\mathcal{T},b_t)$
      \EndIf
    \EndIf
  \EndIf
  \State emit $b_t$ (if any) as current hand box for the gesture detector
\EndWhile
\end{algorithmic}
\end{algorithm}
\vspace{-10pt}

%% file: Sections/5_experimental_setup.tex
\section{Experimental Setup}
\label{sec:exp_setup}
%======================================================================

\subsection{Hardware and Software}

\textbf{Training Multi-GPU Machines:}
All detectors (baseline and synthesized families) are trained on multiple multi-GPU servers, each with 3x NVIDIA RTX~5880-Ada GPUs (48\,GB GDDR6, 14{,}080 CUDA cores each), dual 32-core CPUs, and 512\,GB RAM.
The software stack is \textsc{Ubuntu~22.04}, CUDA~12.2, cuDNN~9.1, Python~3.10, PyTorch~2.5, and Ultralytics~8.3. Lambda Tensorbook equipped with an NVIDIA GeForce RTX~3080\,Ti GPU, an Intel Core i7 CPU.

\textbf{Edge Evaluation Node:}
All ACE, run-time selector, and KF-ROI experiments are run on an NVIDIA Jetson Orin AGX (64\,GB) developer kit with JetPack~6.0 (L4T~36.3, CUDA~12.2, cuDNN~9.x).
We fix the 15\,W power mode and lock CPU and GPU clocks at their maximum frequencies allowed by this TDP.
Videos (DSG-18) are decoded and processed locally on the Orin. Run-time evaluations for dynamic scenarios are carried out on Lambda Tensorbook.

\textbf{Telemetry and Power Logging:}
On Orin we log rail power from \texttt{VIN\_SYS\_5V0} using \texttt{tegrastats} at 50\,ms cadence.
A background daemon \texttt{system\_monitor.py} records a CSV with timestamp, battery percentage (if available), CPU/GPU temperatures, GPU utilization, and rail power every 5\,s.
On desktop sanity runs, we use \texttt{nvidia-smi} with \texttt{--loop-ms=50} to sample board power.

% \subsection{Datasets and Splits}

% All detectors are trained on the full 18-class HaGRID dataset; dataset construction, class mapping, and statistics are detailed in Sec.~\ref{subsec:datasets_ace}.
% We use the standard HaGRID train/validation split and discard corrupted images.
% All run-time profiling and adaptive-selection experiments use the DSG-18 suite with fixed video/JSONL pairs discovered by matching stems. No DSG-18 frames are used for training or fine-tuning.

% \subsection{Training Details}

% Unless stated otherwise, we train each model family (Sec.~\ref{flowchart}) with:

% \begin{itemize}
%   \item Input size 640\(\times\)640, batch size 64.
%   \item Optimizer: SGD with momentum 0.9 or Adam, initial learning rate 0.01, cosine decay, weight decay \(1\times10^{-4}\).
%   \item 200 epochs, early stopping after 30 epochs of no improvement on HaGRID validation \(\mathrm{mAP@0.5}\).
%   \item Standard YOLO augmentations: random horizontal flip, random affine, color jitter, HSV, Mosaic/Copy-Paste.
% \end{itemize}

% For each family, we keep the checkpoint with the best HaGRID validation \(\mathrm{mAP@0.5}\).
% These frozen checkpoints are used in all subsequent ACE profiling, run-time-selector, and KF-ROI experiments; no task- or device-specific fine-tuning is performed.
\vspace{-5pt}
\subsection{Profiling Grid and Metrics}

\textbf{Configuration Grid:}
A configuration is \(x=(m,r,k)\), where \(m\) is a trained detector instance, \(r\in\{160,320,640\}\) is the input resolution, and \(k\in\{1,2,3,6\}\) is the frame stride.
For each \(x\) and for each DSG-18 video/GT pair, we run the \texttt{run\_one\_video} procedure described in Sec.~\ref{subsec:ace-profiling} on the NVIDIA Jetson AGX Orin.

\textbf{Accuracy:}
For every $k$-th frame we keep the highest-confidence prediction, apply a hold-last scheme with exponential confidence decay on skipped frames, and compare against JSONL labels using IoU\(\ge0.5\).
We compute frame-level precision/recall/F1 and event-level F1 based on temporal segments, then form a blended accuracy
\(A(x) = 0.6\,\mathrm{F1}_{\text{event}} + 0.4\,\mathrm{F1}_{\text{frame}}\),
averaged across all DSG-18 videos.

\textbf{Kalman-Gated ROI Parameters:}
For KF-based ROI tracking (Sec.~\ref{subsec:kf-roi}) we use a single-hand Kalman filter over bounding box center and size, ROI scale factor \(s\in[1.6,1.8]\), IoU gate \(\tau\approx 0.5\), and miss budget \(T_{\text{miss}}=8\)–10 frames before full-frame re-acquisition.
We log per-frame \texttt{track\_active} and rail power and post-process these logs to obtain power traces and average $\Delta W$ for ROI vs.\ full-frame baselines.

%% file: Sections/6_results.tex
\section{Results and Analysis}
\begin{figure}[ht]
  \centering
  \includegraphics[width=1.0\linewidth]{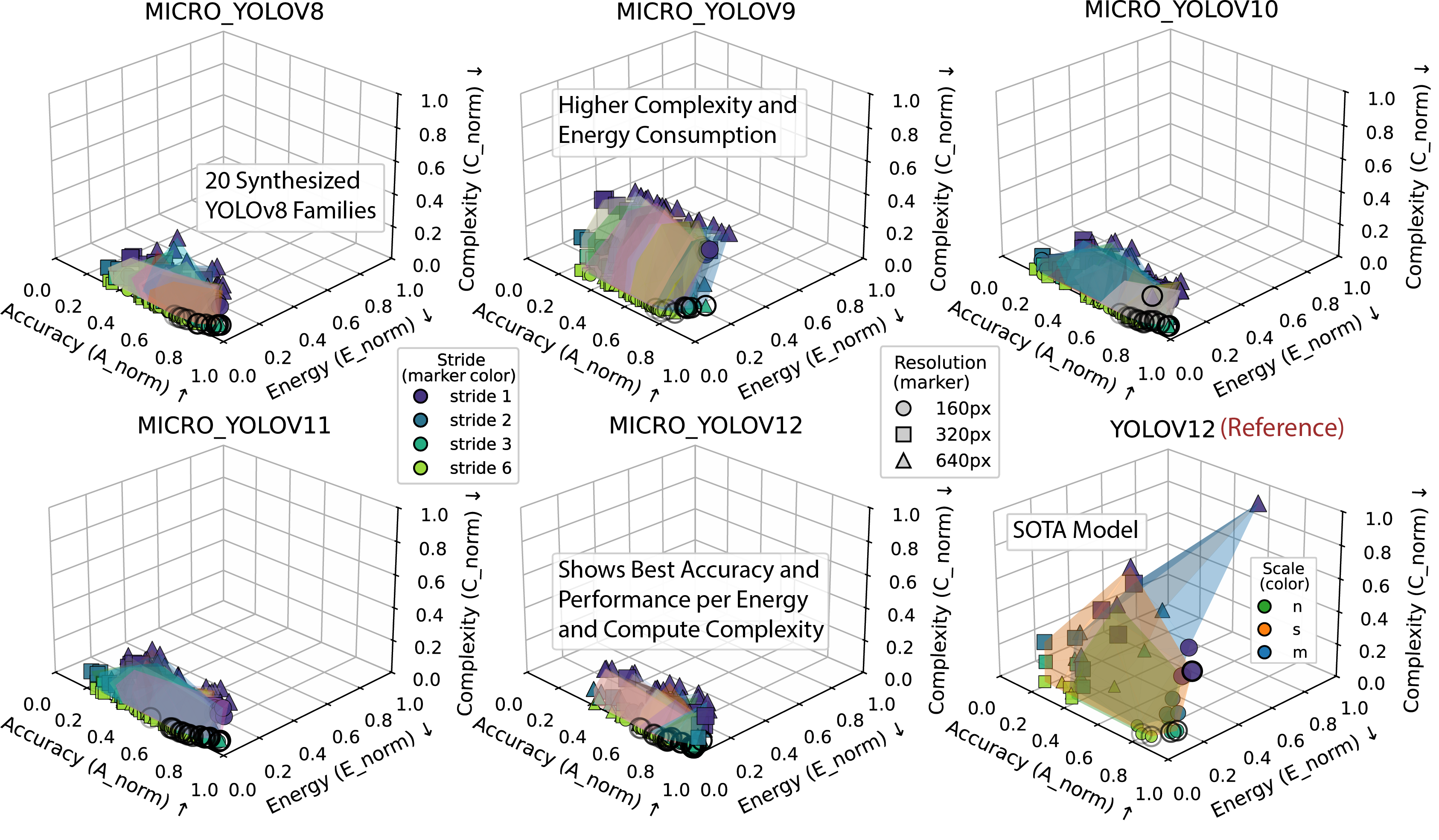}
  \caption{ACE surfaces for micro-scale YOLOv8–12, showing the trade-off between normalized accuracy, complexity, and energy across resolutions and temporal strides; surfaces close to the bottom edge correspond to configurations with lower compute complexity, energy, and high accuracy ACE profiles.}
  \label{fig:surf}
\end{figure}

Fig.~\ref{fig:surf} shows that each synthesized YOLO family forms a smooth ACE trade-off surface: low-resolution, high-stride settings sit in the low-energy/low-complexity corner, while 640~px and low-stride configurations move the surface toward higher accuracy at higher cost. The similar surface shapes across YOLOv8–12 indicate that ACE behaves consistently across families and exposes clear Pareto fronts for the run-time scheduler to exploit.

Table 2 shows that Kalman-gated ROI delivers consistent latency/energy gains with negligible accuracy loss across all three backbones, reducing inference time and energy by roughly 20–50\%. This empirically validates KF-ROI as a lightweight, model-agnostic spatial knob that shifts ACE profiles toward lower complexity and energy, aligning with the design goals of Scale-Gest.

% In Table~\ref{tab:kf_roi}, KF-ROI reduces the average power above idle from $2.21$W to $1.74$W compared to full-frame inference, yielding substantially lower integrated energy over the run. Within the KF-ROI pipeline, tracker ON and OFF states have very similar power, with a small reduction when tracking is active ($\Delta\mu=-0.16$W), indicating that Kalman-based ROI tracking is essentially power-neutral.
\vspace{-5pt}
\begin{table}[ht]
\centering
\scriptsize
\label{tab:kf_roi}
\caption{Runtime and energy–accuracy comparison of full-frame vs.\ KF-ROI across YOLOv12m and micro-scale variants \textit{solar} and \textit{mercury} (v12 solar: $(\alpha,\beta,c_{\max},\text{heads})=(0.25,0.125,320,\text{P3})$, v8 mercury: $(0.18,0.15,192,\text{P3})$).}
\begin{tabular}{lccccccc}
\toprule
Model & Rec. & Prec. & AP$_{50}$ & Infer (ms) & Energy (J) \\
\midrule
YOLOv12m, full-frame (SOTA)
  & \cellcolor{green!15}\textbf{0.996}
  & 0.963
  & \cellcolor{green!15}\textbf{0.993}
  & 65.3
  & 703 \\
YOLOv12m, KF-ROI (x1.8)
  & 0.995
  & \cellcolor{green!15}0.981
  & 0.992
  & 49.2
  & 355 \\
YOLOv12m, KF-ROI (x1.5)
  & 0.989
  & 0.979
  & 0.988
  & \cellcolor{green!15}48.0
  & \cellcolor{green!15}351 \\
\midrule
YOLOv12-solar, full-frame
  & \cellcolor{green!15}0.980
  & 0.975
  & \cellcolor{green!15}0.978
  & 34.7
  & 205 \\
YOLOv12-solar, KF-ROI (x1.8)
  & \cellcolor{green!15}0.980
  & \cellcolor{green!15}\textbf{0.986}
  & \cellcolor{green!15}0.978
  & 27.7
  & 171 \\
YOLOv12-solar, KF-ROI (x1.5)
  & 0.974
  & \cellcolor{green!15}\textbf{0.986}
  & 0.973
  & \cellcolor{green!15}\textbf{26.8}
  & \cellcolor{green!15}\textbf{160} \\
\midrule
YOLOv8-mercury, full-frame
  & \cellcolor{green!15}0.940
  & 0.978
  & \cellcolor{green!15}0.937
  & 34.4
  & 200 \\
YOLOv8-mercury, KF-ROI (x1.8)
  & 0.895
  & 0.980
  & 0.894
  & 27.7
  & 169 \\
YOLOv8-mercury, KF-ROI (x1.5)
  & 0.894
  & \cellcolor{green!15}0.983
  & 0.892
  & \cellcolor{green!15}27.4
  & \cellcolor{green!15}165 \\
\bottomrule
\end{tabular}
\end{table}

% Fig.~\ref{fig:realtest} shows the ACE controller maintains high gesture-recognition F1 and low latency while progressively reducing energy per frame, keeping CPU/GPU temperatures and battery state-of-charge within safe operating ranges. During run-time, the adaptive selector increases the scaling factor $\gamma_C$ while the system latency is increasing, to switch to a lower complexity model which eventually maintains the system latency.

\begin{wrapfigure}[17]{l}{0.24\textwidth}
  \vspace{-15pt}                         
  \setlength{\intextsep}{2pt}       
  \setlength{\columnsep}{-30pt}      
  \centering
  \includegraphics[width=1.2\linewidth]{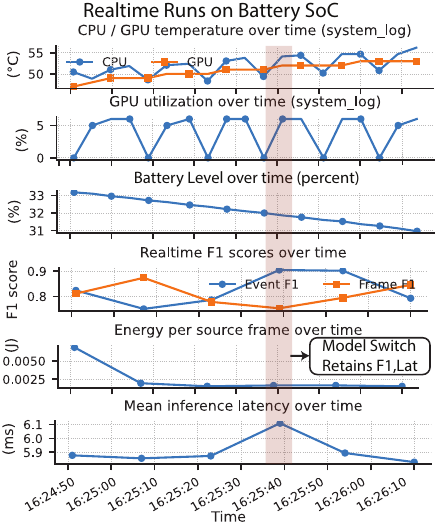}
  % \captionsetup{aboveskip=2pt, belowskip=0pt}
  \caption{Real-time ACE validation on a battery-powered laptop.}
  \label{fig:realtest}
\end{wrapfigure}

Fig.~\ref{fig:realtest} shows that the ACE controller maintains high F1 and low latency while reducing energy per frame and keeping thermals and battery within safe limits. When latency rises, the selector increases the complexity weight $\gamma_C$ to favor lower-complexity models and restore the target frame rate.

\begin{figure}[ht]
  \centering
  \includegraphics[width=1\linewidth]{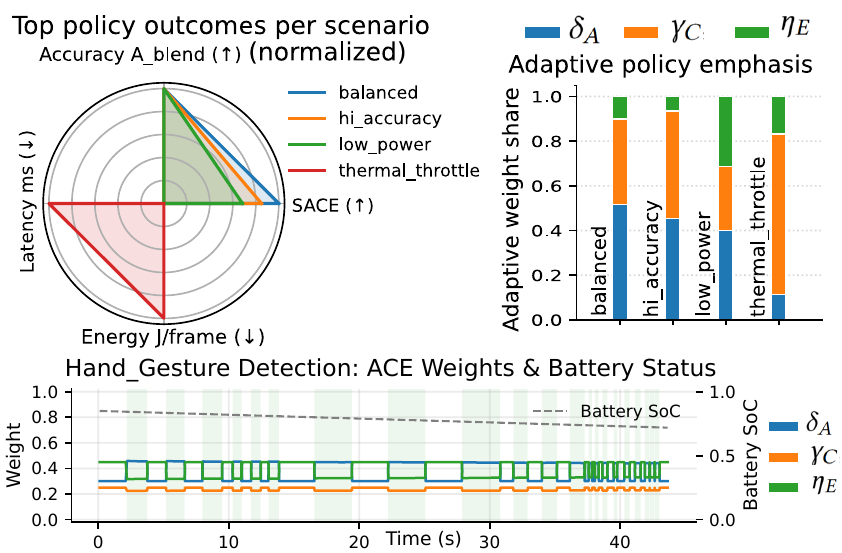}
    \caption{Dynamic ACE-aware Pareto strips for the top five tiers selected under four run-time scenarios.}
    
  \label{fig:radar}
\end{figure}

Across scenarios, Fig.~\ref{fig:radar} shows that the scheduler consistently picks near-Pareto tiers and smoothly trades accuracy for latency and energy as constraints tighten. The balanced and high-accuracy modes favour tiers with slightly higher $A_{\text{blend}}$ at moderate cost in latency and energy, whereas the thermal-throttle modes move to more energy-efficient tiers with only a modest drop in accuracy. The light-green bands mark gesture-active intervals, during which the controller increases the $\delta_A$, temporarily favoring higher-tier models before reverting to efficiency-driven weights in idle segments.
\vspace{-10pt}

\vspace{-10pt}

%% file: Sections/7_conclusion.tex
\section{Conclusion}
This paper introduces Scale-Gest, a run-time adaptive gesture detection framework that targets joint optimization of Accuracy, Complexity, and Energy (ACE) profiles on edge-AI devices. 
We synthesize dense model pipelines, train them on the full 18-class HaGRID dataset, and profile each \textit{model--resolution--stride configuration} on the target SoC using DSG-18. 
At run time, an \emph{Adaptive Selector} combines application-level budgets with live telemetry to derive adaptive ACE weights and rank operating modes, enabling lightweight yet context-aware model selection. 
A Kalman-gated ROI tracker further reduces effective workload by focusing inference on a prediction-driven crop while preserving full-frame accuracy. 
Scale-Gest shows that dense ACE modes, video-based profiling, and lightweight run-time control can systematically meet latency and energy constraints across edge devices. 
Although instantiated for driver hand gestures and YOLO detectors, the methodology is broadly applicable to other on-device vision tasks, model families, and hardware platforms.

\section*{Acknowledgment}
This work was partially supported by the NYUAD Center for Artificial Intelligence and Robotics (CAIR), funded by Tamkeen under the NYUAD Research Institute Award CG010.

\newpage